\documentclass[11pt,a4paper,dvipsnames]{article} %
\usepackage[hyperref]{acl2019}
\usepackage{times}
\usepackage{latexsym}
\usepackage{url}
\usepackage{flushend}

\aclfinalcopy 

\usepackage{caption}
\usepackage{subcaption}
\usepackage{pgfplotstable}
\usepackage{graphicx}
\usepackage{booktabs}
\usepackage{adjustbox}

\usepackage{times}
\usepackage{latexsym}
\usepackage{pgfplots}
\usepackage{amsmath}
\usepackage{rotating}
\usepackage{colortbl}
\definecolor{lcolor}{HTML}{268bd2}
\definecolor{gcolor}{HTML}{c11b17}
\definecolor{wcolor}{HTML}{859900}
\definecolor{ccolor}{HTML}{ffcc00}

\usepackage[utf8]{inputenc}
\usepackage{xargs}                      

\usepackage[colorinlistoftodos,prependcaption,textsize=tiny]{todonotes}
\newcommandx{\unsure}[2][1=]{\todo[linecolor=red,backgroundcolor=red!25,bordercolor=red,#1]{#2}}
\newcommandx{\change}[2][1=]{\todo[linecolor=blue,backgroundcolor=blue!25,bordercolor=blue,#1]{#2}}
\newcommandx{\info}[2][1=]{\todo[linecolor=OliveGreen,backgroundcolor=OliveGreen!25,bordercolor=OliveGreen,#1]{#2}}
\newcommandx{\improvement}[2][1=]{\todo[linecolor=Plum,backgroundcolor=Plum!25,bordercolor=Plum,#1]{#2}}
\newcommandx{\thiswillnotshow}[2][1=]{\todo[disable,#1]{#2}}

\pgfplotstableset{
    color cells/.style={
        col sep=comma,
        string type,
        postproc cell content/.code={%
                \pgfkeysalso{@cell content=\rule{0cm}{2.4ex}\cellcolor{black!##1}\pgfmathtruncatemacro\number{##1}\ifnum\number>50\color{white}\fi##1}%
                },
        columns/x/.style={
            column name={},
            postproc cell content/.code={}
        }
    }
}

\usetikzlibrary{pgfplots.groupplots}
\pgfplotsset{compat=1.11}
\usepgfplotslibrary{ternary}
\usepackage{tikz}
\usetikzlibrary{arrows.meta}
\usepackage{tkz-graph}
\usetikzlibrary{positioning,automata}
\usetikzlibrary{calc,spy,shapes}
\usetikzlibrary{arrows,decorations.pathmorphing,fit,positioning,patterns}
\pgfplotsset{compat=newest}

\title{Blackbox meets blackbox: Representational Similarity and Stability Analysis of Neural Language Models and Brains}

\author{Samira Abnar \quad Lisa Beinborn \quad Rochelle Choenni \quad Willem Zuidema \\ \\
Institute for Logic, Language and Computation \\
  University of Amsterdam \\ \\
  \texttt{\{abnar,l.beinborn\}@uva.nl,rochelle.choenni@student.uva.nl,zuidema@uva.nl} \\
}

\date{}

\begin{document}

\maketitle
\begin{abstract}
In this paper, we define and apply \emph{representational stability analysis} (ReStA), an intuitive way of analyzing neural language models. ReStA is a variant of the popular \emph{representational similarity analysis} (RSA) in cognitive neuroscience. While RSA can be used to compare representations in models, model components, and human brains, ReStA compares instances of the \emph{same} model, while systematically varying single model parameter. 
Using ReStA, we study four recent and successful neural language models, and evaluate how sensitive their internal representations are to the amount of prior context. Using RSA, we perform a systematic study of how similar the representational spaces in the first and second (or higher) layers of these models are to each other and to patterns of activation in the human brain. 
Our results reveal surprisingly strong differences between language models, and give insights into where the \emph{deep} linguistic processing, that integrates information over multiple sentences, is happening in these models. The combination of ReStA and RSA on models and brains allows us to start addressing the important question of what kind of linguistic processes we can hope to observe in fMRI brain imaging data. In particular, our results suggest that the data on story reading from Wehbe et al.\/ (2014) contains a signal of \emph{shallow} linguistic processing, but show no evidence on the more interesting \emph{deep} linguistic processing.
\end{abstract}


\section{Representational Similarity}
Representational similarity analysis (RSA) is a technique which allows us to compare heterogeneous representational spaces~\cite{laakso2000content}. It is very common in cognitive neuroscience because it allows researchers to study the relation between patterns of activation in the brain and representations of stimuli in a computational model \citep{kriegeskorte2008representational}. The key idea is simple: instead of directly trying to map models to brains, we first construct two similarity matrices that record how similar brain responses are to each other for different stimuli, and how similar the computational model's representations for each stimulus are to each other. The representational similarity score is then defined as the similarity (typically: Pearson's correlation) of the two similarity matrices (or equivalently: the similarity of two distance matrices).

RSA can also be applied to deep learning models \cite{laakso2000content,dharmaretnamFyshe18naacl,alvarez2018gromov, wang2018towards,chrupala2019correlating}. In this paper, we present a large-scale study and comparison of both neural language models and fMRI data from brain imaging experiments with human subjects, using RSA. However, 
we extend standard RSA using an approach we call \emph{Representational Stability Analysis} (ReStA). The idea is again simple: we apply RSA to compare instances of the \emph{same} model, while systematically varying a model parameter. 

We focus on a single parameter: the length of the prior context presented to the model.
Varying the amount of context allows us to quantify the degree of context-dependence of different neural language models, and different components of those models. If internal representations are similarly organized regardless of how much additional context is presented to the model, context-dependence is low. If, on the other hand, representations change with each additional amount of context included, context-dependence is high. Using this approach, we find intriguing differences between some recent, successful neural language models (GoogleLM, ELMO, BERT and the Universal Sentence Encoder; Table~\ref{tbl_computational_models}), and between the first and deeper layers of those models.

\newcommand{\multilinecell}[2][c]{%
\begin{tabular}[#1]{@{}c@{}}#2\end{tabular}}
\begin{table*}[btp]
\centering
\begin{adjustbox}{width=\linewidth}
\begin{tabular}{lccrl}
\toprule
\textbf{Model} &\textbf{Objective} & \textbf{Corpus} & \textbf{Rep.Dim.} & \textbf{Architecture} \\ \midrule
GloVe~\cite{pennington2014glove} & \multilinecell{Predicting \\ co-occurrence probabilities} & Wikipedia & 300 & Bag of words\\[0.5cm]
ELMO~\cite{peters2018deep} & \multilinecell{Bidirectional \\ Language Modelling} & 1B benchmark & 1,024 & BiLSTM\\[0.5cm]
GoogleLM~\cite{jozefowicz2016exploring} & Language Modelling & 1B benchmark & 1,024 &  LSTM\\[0.5cm]
UniSentEnc.~\cite{cer2018universal} & \multilinecell{Skip-Thought/Classification} & \multilinecell{Variety of web sources \\ / SNLI} & 512 & Transformer Encoder\\ [0.5cm]
BERT (base)~\cite{devlin2018bert} & \multilinecell{Masked Language Modelling \\ / Next Sent. Pred.} & \multilinecell{BooksCorpus \\ / English Wikipedia} & 768 &  Transformer Encoder\\ [0.5cm]
\bottomrule
\end{tabular}
\end{adjustbox}
\caption{\label{tbl_computational_models} Details of the third party computational models used in this paper, including a brief characterization of the optimization objective, the training corpus, and the dimensionality of representations we extract from them.}
\vspace{-10pt}
\end{table*}

Context-dependence, in turn, gives us a handle on an important question in the research that tries to link neural language models to brain activation: which aspects of language processing in the brain can we hope to observe in fMRI data using NLP and machine learning tools? 


\section{Bridging NLP Models and Neurolinguistics}
An important motivation behind our work is to contribute to answering a big question in computational linguistics: how do we establish a relationship between NLP models and data on the human brain activation while they process language? Pioneering work of \citet{mitchell2008} showed that techniques from distributional semantics could be used to predict and decode brain activation. In the decade since that paper, many efforts have been reported using brain data to evaluate computational models, or using NLP models to build predictive models of the human brain, or both~\cite{Murphy2012,wehbe2014,ruan2016exploring,Sogaard2016,xu2016,fyshe2014interpretable,Bingel2016,Bulat2017, Abnar2017, pereira2018toward, huth2016natural}. 

Most of that work is focused on lexical representations, reporting promising results for concrete nouns, presented in isolation. More recently researchers have tried to adapt the methodology to address words in context, in sentence and story processing tasks. 
\citet{pereira2018toward}, for instance, used a bag of words model of sentence meaning to decode sentences from brain activation. \citet{Wehbe2014AligningCS, Qian2016} use the internal states of LSTMs trained for language modelling for encoding. \citet{Jain327601} report that the higher layers of the LSTM are better at predicting the activation of brain regions that are known for higher level language functions (a finding seemingly at odds with results from section\ref{sec:exp:rsa}).

In this effort, however, we run into a number of major conceptual, methodological and technical challenges. Most importantly: how do we determine what we are really observing in the brain data? Are we really seeing signatures of linguistic processes, or just neural correlates of general cognitive processes evoked by a correct understanding of the linguistic input? How do we adequately control for alternative explanations of the observed correlations? And how do we deal with the intricate temporal dynamics and the overwhelmingly high dimensionality of the brain, and the very indirect, delayed and/or coarse measurements that neuroimaging gives us of the processes in the brain? Merely demonstrating a correlation between two black boxes is clearly not sufficient. 

\begin{figure*}[t!]
    \centering
    \includegraphics[width=0.7\textwidth]{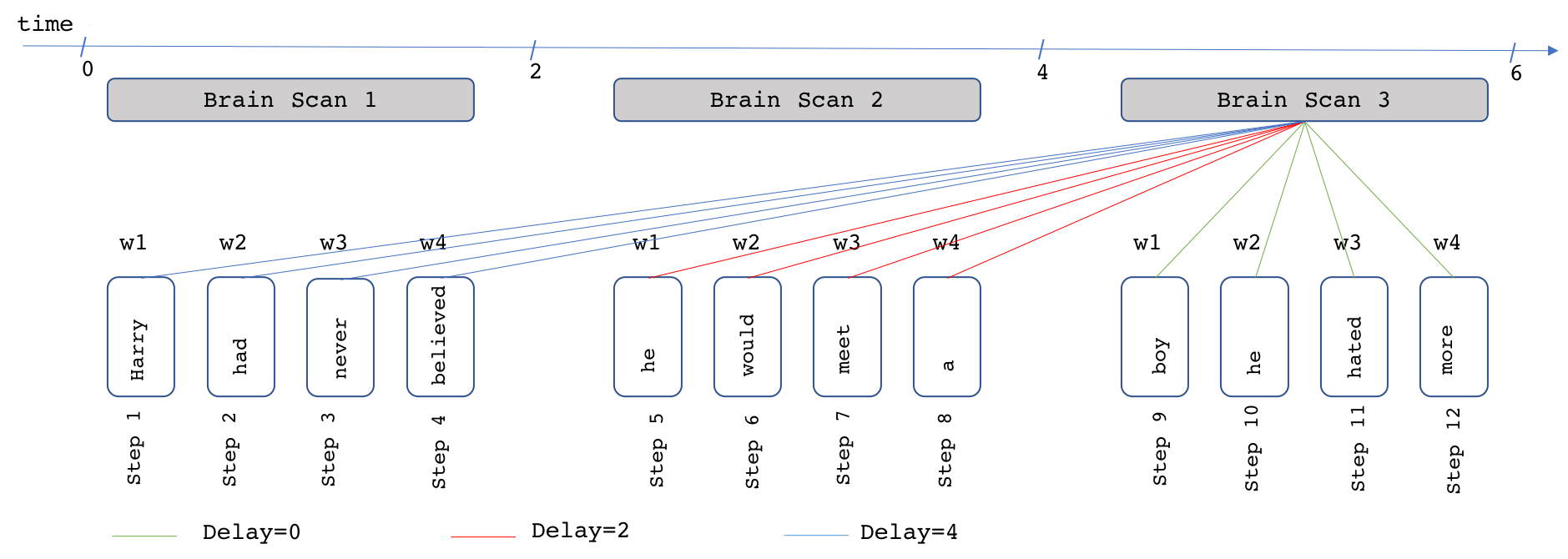}
    \caption{Alignment of the words in the story and the brain vectors. Each fMRI scan lasts for 2 seconds during which the subject is reading four words sequentially.
    Delay is the amount of time in seconds between the time the first of the four word is shown to the subject and when the fMRI scan is started to be taken. 
    \label{fig:diag}
    }
\end{figure*}
\begin{table}[b]
    \centering
    \begin{adjustbox}{max width=\columnwidth}
    \begin{tabular}{cccccc}
        \toprule
         Block & Words & Unique words & Sentences & Sent Length & Scans \\
         \midrule
         1 & 1583 & 553 & 115 & 11 & 326\\
         2 & 1711 & 560 & 163 & 8 & 338 \\
         3 & 1411 & 461 & 134 & 8 & 265\\
         4 & 1853 & 583 & 177 & 8 & 366\\
         \bottomrule
    \end{tabular}
    \end{adjustbox}
    \caption{Statistics of the Harry Potter dataset.}
    \label{tab:harrystat}
\vspace{-10pt}
\end{table}

\begin{figure}
    \centering
    \begin{subfigure}[b]{\columnwidth}
    \includegraphics[width=\textwidth-15pt,clip]{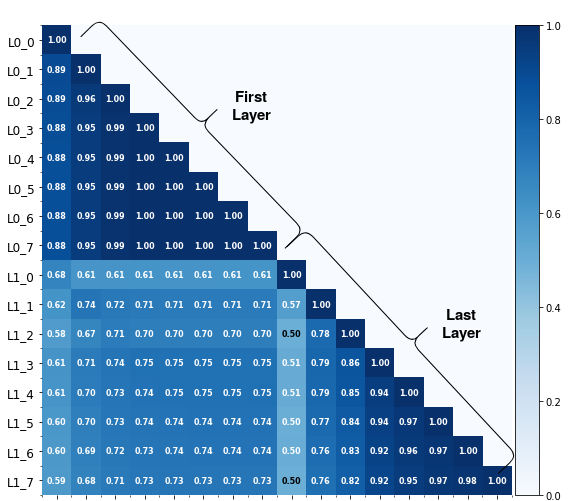}
    \vspace{-2pt}
    \caption{GoogleLM\label{fig:google_lm_sim}}
    \end{subfigure}
    \begin{subfigure}[b]{\columnwidth}
    \includegraphics[width=\textwidth-15pt, clip]{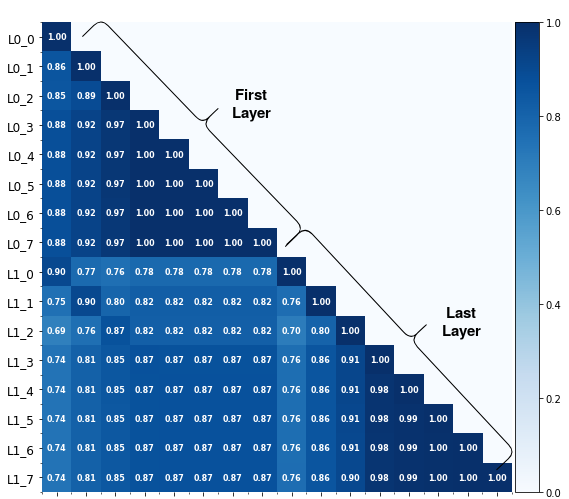}
    \vspace{-2pt}
    \caption{ELMO\label{fig:elmo_lm_sim}}
    \end{subfigure}
     \begin{subfigure}[b]{\columnwidth}
    \includegraphics[width=\textwidth-15pt, clip]{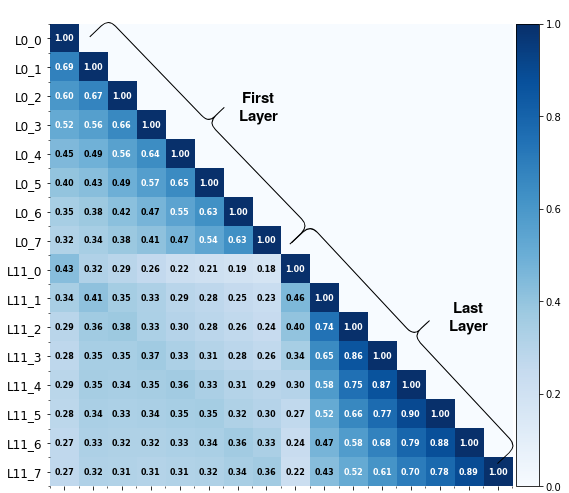}
    \vspace{-2pt}
    \caption{BERT\label{fig:bert_lm_sim}}
    \end{subfigure}
    \caption{RSA between different layers of each model given different context length in terms of number of previous sentences over the story words. In these plots, for example \texttt{L1\_c3} means representation from layer 1, when the context length is 3 sentences including the current sentence. When $c=0$, the model only sees the current words and when $c=1$ the model sees current sentence up to the target word. Here darker means more similar. The values are averaged over the four story blocks and the standard deviation of all the values across the four blocks are below $0.002$.
    \label{fig:lm_sim}}
\end{figure}

We argue that experiments to find the model best correlated with brain activations should be accompanied by efforts for interpreting the internal representations and operations of the models. Applying ReStA for the prior context parameter gives us a way to roughly characterize the \emph{depth} of linguistic processing in different language models and different components of these models. If a model component only tracks the lexical semantics of the current word, the representations it forms should not be sensitive to the amount of prior context. On the other hand, If a model component tracks long-distance syntactic dependencies, semantic polarity, named entities, topics or story arcs, resolves anaphora or builds up situation models, its representations will be different whenever different amounts of prior context are available. Hence, in this paper, we will interpret context-dependence as an imperfect but useful signature of deep linguistic processing.


\section{Models and Data}
In this section, we explain the language encoding models we study in our experiments and the dataset from which we get the language stimuli and their corresponding brain data.  

\subsection{Neural Language Models}
We study language models with different architectures trained with different objective functions (see Table~\ref{tbl_computational_models}).
As a word level embedding model, we use GloVe~\cite{pennington2014glove}. We consider a sentence as a bag of words and take the average of the GloVe embeddings of its individual words.

We employ two high performing LSTM based language models: ELMO~\cite{peters2018deep} and GoogleLM~\cite{jozefowicz2016exploring}.
Both of these models have two LSTM layers; however, ELMO uses bidirectional LSTM layers, whereas in the GoogleLM the LSTM layers are uni-directional. From these models, we take the internal states of each of the LSTM layers as two different representation spaces.

In our comparisons, we also use BERT and the Universal Sentence Encoder (UniSentEnc), as Transformer based models. BERT is trained on masked language modelling and next sentence prediction tasks~\cite{devlin2018bert} while the Universal Sentence Encoder is trained on a different objective than language modelling. The parameters of this model are optimized with respect to different language tasks such that it can better encode the meaning of complete sentences. 
These two models do not have the recurrent inductive bias of LSTMs, and hence the representations they learn can be completely different. 

To study how and where the models integrate information over time, we modify the amount of context provided to the models to obtain the contextualized word representations. We do this at the sentence level. Thus, for the context length of $0$, we only feed the target words to the models; For context length $1$ we feed all the previous words in the current sentence to the models. For context length $i$ where $i > 1$, in addition to the current sentence we feed all the words in the last $i$ sentences. We operate on the sentence level to feed the model with independently meaningful pieces of text. 

From prior work, we expect a relation between the depth of the layers and the level of abstraction of their representations. We study this intuition here empirically by analyzing the different layers of the models, and we focus on the first and last layers. Note that the last layer corresponds to the second layer for the LSTM architectures, but to the 12th layer for Bert.
\subsection{Brain Data}
We compare the representations of our model to human brain activations captured while reading a story. We use the dataset by~\citep{wehbe2014} which consists of the fMRI scans of 8 participants reading chapter 9 of \textit{Harry Potter and the Sorcerer's stone} \cite{Rowling}.\footnote{The data is available at \url{http://www.cs.cmu.edu/~fmri/plosone/}. Further information on the pre-processing steps is described in the supplementary material.} 

The story was presented to the participants word by word on a screen in four continuous blocks.\footnote{The story chapter is split into four almost equal length blocks, each reflecting approximately 12 minutes of measurements. Each block is presented to the participant in one continuous trial, and experimental blocks are separated by pauses for the subjects.} 
Each word was displayed for 0.5 seconds and an fMRI scan was taken every 2 seconds. Figure \ref{fig:diag} visualizes an example for the beginning of the chapter. More detailed statistical information about the stimuli can be found in Table~\ref{tab:harrystat}.

\paragraph{Brain Regions}
The fMRI data contains activation values for approximately 40,000 voxels per scan, each reflecting the oxygen usage (the ``BOLD response'') in approximately $3mm^3$ of brain tissue. 
To obtain the brain representations, we flatten the 3D fMRI images into vectors thereby ignoring the spatial relationships between the voxels. We do this either for the whole brain, or for specific regions separately.
Not all of the scanned voxels are related to language processing, but the changes in activity might be associated with other cognitive processes like, for example, the noise perception in the scanner.
A common reduction method is to restrict the brain response to voxels that fall within a pre-selected set of regions.
In our analysis, we only include the voxels from the top $k$ regions that are most similar across different subjects given the same stimuli. We heuristically set the value of $k$ to $16$ based on the distribution of the similarity scores.\footnote{We sort the brain regions based on their cross-subject similarities for different stimuli and pick a threshold value based when there is a relatively big jump in the similarity scores.}

\paragraph{Delay} An important point to consider when dealing with fMRI data is the hemodynamic response delay: from the time neurons start firing, it takes $4$ to $6$ seconds until the Bold response reaches its peak~\cite{buckner1998event}. This means that from the time a stimulus is presented to a subject, it takes approximately $5$ seconds before we can observe its response in the fMRI scan of the brain. 
We account for this delay by varying the alignment between stimuli and scans. If we apply a delay of $0$ seconds, scan 3 in the example would be applied to the sequence \textit{boy he hated more}, Figure \ref{fig:diag}. With a delay of $2$ seconds, it is aligned to the previous stimulus \textit{he would meet a} and a delay of $4$ would result in alignment with \textit{Harry had never believed}.


\begin{figure}[t]
    \centering
    \includegraphics[width=\columnwidth, clip]{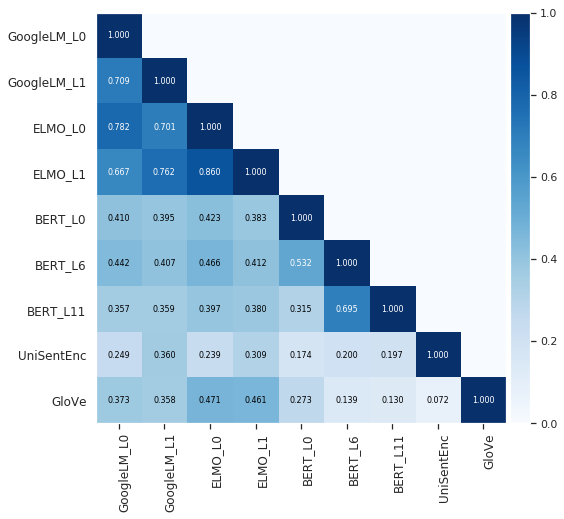}
    \caption{RSA across models
    \label{fig:cross_lm_sim}}
\end{figure}
\pgfplotstableread{
|C| GoogleLM(L0) GoogleLM(L1) Elmo(L0) Elmo(L1) Bert(L0) Bert(L11) UniSentEnc GloVe 
0  0.890723  0.508319  0.840976  0.723882  0.696842  0.446847  0.174503  0.996179
1  0.966706  0.768874  0.894703  0.796853  0.590288  0.722427  0.588583  0.996994
2  0.989405  0.867519  0.975085  0.913422  0.627383  0.858998  0.691391  0.994691
3  0.998365  0.944986  0.997083  0.981142  0.566678  0.860661  0.758843  0.995801
4  0.999665  0.970860  0.999057  0.993766  0.546098  0.869912  0.847928  0.997577
5  0.999928  0.980200  0.999801  0.996369  0.574377  0.884567  0.839262  0.995433
6  0.999974  0.984828  0.999888  0.998008  0.584614  0.901597  0.855207  0.997322
}\tableone

\pgfplotstableread{
ContextLength	GoogleLM(L0->L1)	Elmo(L0->L1)	Bert(L0->L1) Bert(L0->L11)
0  0.657877  0.892764  0.396767  0.567703
1  0.719601  0.883588  0.346806  0.594112
2  0.691488  0.848529  0.300975  0.562313
3  0.728551  0.852779  0.285274  0.520578
4  0.729289  0.852131  0.291903  0.475020
5  0.718492  0.849923  0.285820  0.473008
6  0.715591  0.849376  0.299730  0.507758
7  0.710471  0.850560  0.311346  0.551703
}\tabletwo

\pgfplotstableread{
|C| GoogleLM(L0) GoogleLM(L1) Elmo(L0) Elmo(L1) Bert(L0) Bert(L11) UniSentEnc GloVe 
0	0.075983	0.260555	0.053727	0.072971	-0.106554	0.275580	0.414081	0.000815
1	0.022699	0.098645	0.080382	0.116570	0.037095	0.136571	0.102808	-0.002303
2	0.008961	0.077467	0.021998	0.067720	-0.060706	0.001663	0.067452	0.001110
3	0.001299	0.025874	0.001974	0.012624	-0.020579	0.009251	0.089085	0.001777
4	0.000264	0.009340	0.000744	0.002603	0.028279	0.014655	-0.008666	-0.002145
5	0.000046	0.004628	0.000087	0.001639	0.010236	0.017030	0.015945	0.001889
}\tablethree

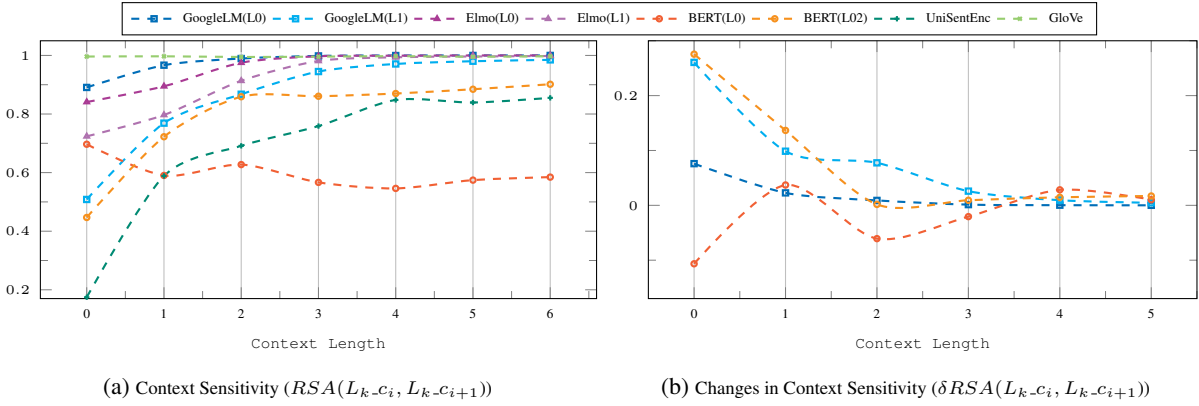
\begin{figure*}
\centering
\begin{subfigure}[t]{0.5\textwidth}
\usetikzlibrary{patterns}
\begin{tikzpicture}
\begin{axis}[
width=\textwidth+0.9cm,
height=5cm,
xmajorgrids,
minor tick num=1,
xlabel={\texttt{Context Length}},
xtick=data,
ymin=0.17, 
ymax=1.05,
xtick={0,1,2,3,4,5,6,7},
xticklabels={0,1,2,3,4,5,6,7},
tick label style = {font=\fontsize{5}{6}\selectfont,         
        /pgf/number format/fixed,
        /pgf/number format/precision=3},
label style = {font=\fontsize{6}{7}\selectfont, yshift=0.01ex},
legend style={at={(1.9,1.15),font=\fontsize{5}{6}\selectfont}
  ,legend columns=8
  },
]
\addplot [thick, NavyBlue, dashed, mark=square,mark options={scale=0.5, solid},smooth] table[x index=0, y index=1]{\tableone};

\addplot [thick, Cyan, dashed, mark=square,mark options={scale=0.5, solid},smooth] table[x index=0,y index=2] {\tableone};

\addplot [thick, Mulberry, dashed , mark=triangle,mark options={scale=0.5, solid},smooth] table[x index=0,y index=3] {\tableone};

\addplot [thick, Orchid, dashed, mark=triangle,mark options={scale=0.5, solid},smooth] table[x index=0,y index=4] {\tableone};

\addplot [thick, RedOrange, dashed, mark=o,mark options={scale=0.5, solid},smooth] table[x index=0, y index=5]{\tableone};

\addplot [thick, BurntOrange, dashed, mark=o,mark options={scale=0.5, solid},smooth] table[x index=0,y index=6] {\tableone};

\addplot [thick, PineGreen, dashed, mark=+,mark options={scale=0.5, solid},smooth] table[x index=0,y index=7] {\tableone};

\addplot [thick, YellowGreen, dashed, mark=x,mark options={scale=0.5, solid},smooth] table[x index=0,y index=8] {\tableone};

\legend{GoogleLM(L0), GoogleLM(L1), Elmo(L0), Elmo(L1), BERT(L0), BERT(L02), UniSentEnc, GloVe}
\end{axis}
\end{tikzpicture}
\caption{\fontsize{7}{8}\selectfont{Context Sensitivity ($RSA(L_k\_c_i, L_k\_c_{i+1}$))
\label{fig:context_sim}}
}
\end{subfigure}%
\begin{subfigure}[t]{0.5\textwidth}
\usetikzlibrary{patterns}
\begin{tikzpicture}
\begin{axis}[
width=\textwidth+0.8cm,
height=5cm,
xmajorgrids,
minor tick num=1,
xlabel={\texttt{Context Length}},
xtick=data,
ymin=-0.17, 
ymax=0.3,
xtick={0,1,2,3,4,5,6,7},
xticklabels={0,1,2,3,4,5,6,7},
tick label style = {font=\fontsize{5}{6}\selectfont,         
        /pgf/number format/fixed,
        /pgf/number format/precision=3},
label style = {font=\fontsize{6}{7}\selectfont, yshift=0.01ex},
]
\addplot [thick, NavyBlue, dashed, mark=square,mark options={scale=0.5, solid},smooth] table[x index=0, y index=1]{\tablethree};

\addplot [thick, Cyan, dashed, mark=square,mark options={scale=0.5, solid},smooth] table[x index=0,y index=2] {\tablethree};

\addplot [thick, RedOrange, dashed, mark=o,mark options={scale=0.5, solid},smooth] table[x index=0, y index=5]{\tablethree};

\addplot [thick, BurntOrange, dashed, mark=o,mark options={scale=0.5, solid},smooth] table[x index=0,y index=6] {\tablethree};

\end{axis}
\end{tikzpicture}
\caption{\fontsize{7}{8}\selectfont{Changes in Context Sensitivity ($\delta RSA(L_k\_c_i, L_k\_c_{i+1}$))
\label{fig:delta_context_sim}}
}
\end{subfigure}
\caption{Changes in RSA by increasing context length. (a) Shows how the amount of difference in the representational spaces changes by increasing the context length. (b) Shows for all models that we study, regardless of whether and how much their representations change by increasing context length, the amount of difference becomes almost constant after context length of 3 sentences. Note that in (b), we have scaled the plot and removed some of the models to increase the readability.
\label{fig:summ_cross_lm}}
\end{figure*}

\begin{figure}
\centering
\usetikzlibrary{patterns}
\begin{tikzpicture}
\begin{axis}[
width=\textwidth+0.9cm,
height=3.5cm,
xmajorgrids,
minor tick num=1,
xlabel={\texttt{Context Length}},
xtick=data,
ymin=0.20, width=\columnwidth,
ymax=1.,
xtick={0,1,2,3,4,5,6,7},
xticklabels={0,1,2,3,4,5,6,7},
tick label style = {font=\fontsize{5}{6}\selectfont,         
        /pgf/number format/fixed,
        /pgf/number format/precision=3},
label style = {font=\fontsize{6}{7}\selectfont, yshift=0.01ex},
legend style={at={(0.87,1.40),font=\fontsize{5}{6}\selectfont}
  ,legend columns=2
  },
]
\addplot [thick, NavyBlue, dashed, mark=*,mark options={scale=0.5, solid},smooth] table[x index=0, y index=1]{\tabletwo};

\addplot [thick, Mulberry, dashed, mark=*,mark options={scale=0.5, solid},smooth] table[x index=0,y index=2] {\tabletwo};

\addplot [thick, RedOrange, dashed, mark=*,mark options={scale=0.5, solid},smooth] table[x index=0,y index=3] {\tabletwo};

\addplot [thick, BurntOrange, dashed, mark=*,mark options={scale=0.5, solid},smooth] table[x index=0,y index=4] {\tabletwo};
\legend{GoogleLM(L0$\rightarrow$L1), Elmo(L0$\rightarrow$L1), BERT(L0$\rightarrow$L02),
BERT(L0$\rightarrow$L1)}
\end{axis}
\end{tikzpicture}
\caption{Layer similarities ($RSA(L_k\_c_i, L_{k+1}\_c_i$). Here we show how increasing context length affects the similarity between different layers of the models.)
\label{fig:layer_diff}}
\vspace{-15pt}
\end{figure}
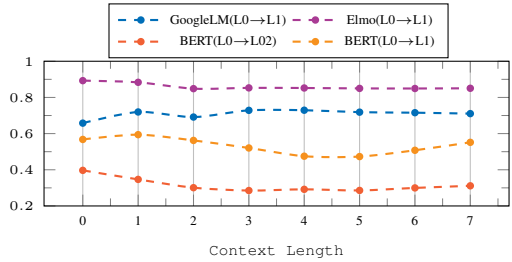

\section{Analyzing Neural Language Models}
In this section, we present the results of applying ReStA, Representational Stability Analysis, to three different language encoding models, GoogleLM, ELMO and BERT. We investigate what type of information is captured in the learned representations without making any explicit assumptions.
Next, we apply standard RSA to, first, investigate the relations between different components of the language encoding models, and second to study the alignment of these components with the activity patterns in the human brain.\footnote{We made the code that reproduces all the experiments publicly available at ~~\{\url{https://github.com/samiraabnar/Bridge}\}}

\subsection{Representational Stability Analysis}
We define the \emph{Representational Stability} as the similarity between the representations obtained from a model, when a single condition is changed, i.e. increase in context length. We use RSA to measure the similarity between the representational spaces. 
And to compute \emph{RSA} we use \emph{cosine} similarity to measure the intra-space similarities and use \emph{Pearson} correlation to quantify the similarities across representational spaces.

In Figure~\ref{fig:lm_sim} the representations of the different layers given different context lengths are compared for GoogleLM, ELMO and BERT. The values under the diagonal of these plots indicate the ReStA when the varying condition is context length. This is measured as $RSA(L_{k-c_i}, L_{k-c_j})$, where $k$ is the layer id and $c_i$ and $c_j$ are different conditions which in this case indicate different context lengths. We have depicted the trends of how the ReStA changes for different context length in Figures~\ref{fig:context_sim} and \ref{fig:delta_context_sim}.

\paragraph{Effect of depth}
As we can see in Figure~\ref{fig:lm_sim} and more clearly in Figure~\ref{fig:layer_diff}, for the LSTM based models, we observe a higher degree of similarity between the two layers ($\sim0.75$ and $\sim0.80$) compared to BERT ($\sim0.35$). This can be partly explained by the higher number of layers in BERT, i.e the first and last layer are further apart.
Moreover, the relation between the first and last layers is almost the same for all context lengths and for all these three models the two layers are most similar when provided with the same amount of context.
\paragraph{Context sensitivity}
Next, we analyse the sensitivity of different layers of each model to context length. In Figures \ref{fig:context_sim} and \ref{fig:lm_sim}, we see that for both LSTM based models, GoogleLM and ELMO, the first layer, $L0$, is less sensitive to the changes in the context length compared to the last layer, $L1$, i.e. the representations are not affected anymore by increasing the context length to more than $3$ sentences.
A hierarchical encoding mechanism, where the first layer is responsible for encoding the local context and the second(last) layer is encoding more global information, can justify these results.

We can see in Figure \ref{fig:context_sim}, that the sensitivity to the context length is more significant in the Transformer based models compared to LSTM based models. In these models, the difference in the representations at different context lengths does not fade away as the context length increases but the rate of the changes becomes constant. 
As illustrated in Figures \ref{fig:context_sim} and \ref{fig:bert_lm_sim} we observe that in BERT, regardless of the current context length, adding more context leads to different representations. In addition, in this model, the representations from the first layer, $L0$ are more context-dependent than those from the last layer, $L11$.
Since in self-attention layers, there is a direct connection between the representations at different positions, the higher degree of sensitivity to context length is not surprising. This is evidence that, for computing the representations of each position in the input, the representations from all positions, no matter how far they are, are in fact taken into account. We speculate that the last layer of BERT is less sensitive to context could be that in higher layers, the representations correspond to more abstract meanings, and the representational space becomes denser than the lower layers.

\subsection{RSA across Models}
In the second step, we study whether the computational models have learned inherently different representational spaces.
According to representational similarity scores, among the models that we study, shown in Figure \ref{fig:cross_lm_sim}, UniSentEnc seems to learn very different representations from ELMO, GoogleLM and BERT.
While BERT and UniSentEnc are both Transformer based models, the representational space of BERT is more similar to the representations from ELMO and GoogleLM that are LSTM based models. This can be due to the fact that ELMO, GoogleLM and BERT are trained with language modelling objectives, while UniSentEnc is trained on skip-thought and classification tasks and this could indicate the effect of the training objective on the representational spaces. 


\section{The Relation between the Models and the Activity Patterns in Human Brains}
\label{sec:exp:rsa}

\begin{figure*}[t]
\centering
\begin{tikzpicture}
\pgfkeys{
    /pgf/number format/fixed,
}
\begin{axis}[
    width= \textwidth,
    height=5cm, 
    enlarge y limits=0.05,
    enlarge x limits=0.1,
    ybar=1pt,
    bar width= 4.8pt,
    symbolic x coords ={GoogleLM(L0),GoogleLM(L1),ELMO(L0),ELMO(L1),BERT(L0),BERT(L11),UniSentEnc,GloVe},
    xtick=data,
    ticklabel style = {font=\fontsize{8}{9}\selectfont},
    ymin=0.08, 
    ymax=0.4,
    label style = {font=\fontsize{6}{7}\selectfont, yshift=0.5ex},
    area legend,
    legend style={legend image post style={xscale=0.3}, at={(0.85,1.18),font=\fontsize{7}{8}\selectfont},
    legend columns=8,
    },
    tick label style={font=\tiny},
    y label style={at={(axis description cs:-0.08,.5)},},
    ]
    \addplot[fill=NavyBlue!40!Yellow, draw=blue!50!black, pattern color = lcolor!20!lcolor,
    error bars/.cd,
     y dir=both,y explicit relative,
    ] table[x=Model,y index=1, y error index=9] \tablethree; 
    
    \addplot[fill=NavyBlue!50!Yellow, draw=blue!50!black, pattern color = lcolor!20!lcolor,
    error bars/.cd,
     y dir=both,y explicit relative,
    ] table[x=Model,y index=2, y error index=10] \tablethree; 
    
     \addplot[fill=NavyBlue!60!Yellow, draw=blue!50!black, pattern color = lcolor!20!lcolor,
     error bars/.cd,
     y dir=both,y explicit relative,
    ] table[x=Model,y index=3, y error index=11] \tablethree; 
    
     \addplot+[fill=NavyBlue!70!Yellow, draw=blue!50!black, pattern color = lcolor!20!lcolor,
     error bars/.cd,
     y dir=both,y explicit relative,
    ] table[x=Model,y index=4, y error index=12] \tablethree; 
    
    \addplot+[fill=NavyBlue!80!Yellow, draw=blue!50!black, pattern color = lcolor!20!lcolor,
     error bars/.cd,
     y dir=both,y explicit relative,
    ] table[x=Model,y index=5, y error index=13] \tablethree; 
    
     \addplot+[fill=NavyBlue!80!Yellow, draw=blue!50!black, pattern color = lcolor!20!lcolor,
     error bars/.cd,
     y dir=both,y explicit relative,
    ] table[x=Model,y index=6, y error index=14] \tablethree; 
    
     \addplot+[fill=NavyBlue!80!Yellow, draw=blue!50!black, pattern color = lcolor!20!lcolor,
     error bars/.cd,
     y dir=both,y explicit relative,
    ] table[x=Model,y index=7, y error index=15] \tablethree; 
    
     \addplot+[fill=NavyBlue!80!Yellow, draw=blue!50!black, pattern color = lcolor!20!lcolor,
     error bars/.cd,
     y dir=both,y explicit relative,
    ] table[x=Model,y index=8, y error index=16] \tablethree; 
    
    \legend{Delay=-4, Delay=-2,Delay=0,Delay=2,Delay=4,Delay=6, Delay=8, Delay=10}
   
\end{axis}
\end{tikzpicture}
\vspace{-5pt}
\caption{Representational similarity of the models and brains averaged over all subjects and the four blocks at different time delays after the human subjects have read the target words, when the context provided to the models is three sentences. Here the delay is increasing from left to right and the error bars indicate the standard deviation across different blocks.}
\label{fig:brain_sim_delay}
\end{figure*}
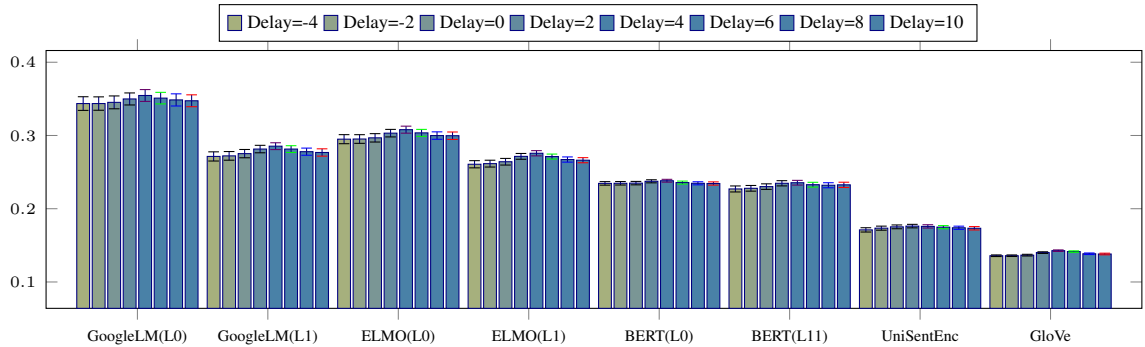

\begin{figure}
\centering
\usetikzlibrary{patterns}
\begin{tikzpicture}
\begin{axis}[
width=\columnwidth+0.9cm,
height=4cm,
xmajorgrids,
minor tick num=1,
xlabel={\texttt{Context Length}},
xtick=data,
ymin=0.13, 
ymax=0.4,
xtick={0,1,2,3,4,5,6,7},
xticklabels={0,1,2,3,4,5,6,7},
tick label style = {font=\fontsize{5}{6}\selectfont,                  /pgf/number format/fixed,         /pgf/number format/precision=3},
label style = {font=\fontsize{6}{7}\selectfont, yshift=0.5ex},
legend style={at={(1.0,1.32),font=\fontsize{5}{6}\selectfont}
  ,legend columns=4
  },
]
\addplot [thick, NavyBlue, dashed, mark=square,mark options={scale=0.5, solid}] table[x index=0, y index=1]{\tablethree};

\addplot [thick, Cyan, dashed, mark=square,mark options={scale=0.5, solid}] table[x index=0,y index=2] {\tablethree};

\addplot [thick, Mulberry, dashed , mark=triangle,mark options={scale=0.5, solid}] table[x index=0,y index=3] {\tablethree};

\addplot [thick, Orchid, dashed, mark=triangle,mark options={scale=0.5, solid}] table[x index=0,y index=4] {\tablethree};

\addplot [thick, RedOrange, dashed, mark=o,mark options={scale=0.5, solid}] table[x index=0, y index=5]{\tablethree};

\addplot [thick, BurntOrange, dashed, mark=o,mark options={scale=0.5, solid}] table[x index=0,y index=6] {\tablethree};

\addplot [thick, PineGreen, dashed, mark=+,mark options={scale=0.5, solid}] table[x index=0,y index=7] {\tablethree};

\addplot [thick, YellowGreen, dashed, mark=x,mark options={scale=0.5, solid}] table[x index=0,y index=8] {\tablethree};

\legend{GoogleLM(L0), GoogleLM(L1), Elmo(L0), Elmo(L1), Bert(L0), Bert(L11), UniSentEnc, GloVe}
\end{axis}
\end{tikzpicture}
\caption{Similarity of the representations from different layers of different models, given different amount of context with brain representations, averaged over all subjects. Note that the average RSA of brains of different human subjects is about 0.55}
\label{fig:avg_modeltobrain}
\end{figure}
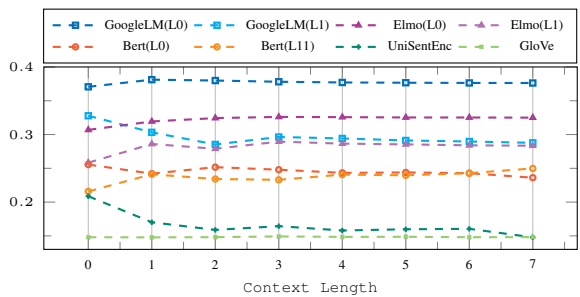

Figure \ref{fig:avg_modeltobrain} shows the similarity of different computational representation spaces with brain representations, with respect to different amounts of context provided to the models, averaged over all human subjects.
Due to the hemodynamic response delay, we expect to see the peak in
similarities after about 4s delay. As we can see in Figure \ref{fig:brain_sim_delay}, the highest RSA for all models is at $Delay=4s$, the ranking of the models based on their similarities with brain representations is the same for all amounts of delay.
Interestingly, the performances of these models on the NLP tasks are not correlated with their similarity with the brain representations (but note the overall low correlations). 
The representations learned by LSTM based models are most similar to the brain data, and for both ELMO and GoogleLM the representations from lower layers, $L0$, have higher similarity scores compared to the higher layers, $L1$.
Interestingly, for UniSentEnc, BERT($L11$) and also GoogleLM($L1$), increasing the context length, which usually boosts the performance of language encoding models in language understanding tasks~\citep{wang2015larger}, leads to lower similarity with brain representations. 
It seems that the way these models integrate the context information, pushes the representation further away from the brain representations.
This could mean:
(1) These models are doing fairly well at encoding the local context, but not at a more global level.
Alternatively, (2) The information about the more global aspects of the meaning is not encoded in the brain representations.

\pgfplotstableread{
Delay	GoogleLM(L0)	GoogleLM(L1)	Elmo(L0)	Elmo(L1) BERT(L0)	BERT(L11) UniSentEnc GloVe
0  0.345213  0.275228  0.296787  0.264157  0.234949  0.230196  0.175264  0.136625
2  0.349875  0.281469  0.303175  0.271473  0.237449  0.234737  0.176276  0.140097
4  0.354602  0.285399  0.307938  0.275734  0.238211  0.235501  0.175887  0.142728
6  0.351002  0.281457  0.303565  0.271237  0.235698  0.232809  0.174843  0.141485
8  0.348563  0.277889  0.299885  0.267168  0.234726  0.232059  0.174002  0.138522    
}\tabletwo

\pgfplotstableread{
Model           Delay=6     Delay=4     Delay=2     Delay=0
GoogleLM(L0)    0.3775870285115762  0.3812284259692499  	0.37658243835458194 0.3730588640988145
GoogleLM(L1)    0.29895023725970554    0.3032681346108651        0.2988682648688573 0.29388291720525195
Elmo(L0)        0.31632012736785614 0.3193476497203331     0.3136933609951579  0.3083983104688391
Elmo(L1)        0.282962040350316  0.2859980247689471     0.27971861626651834    0.27412382969628624
BERT(L0)        0.23882216055801714    0.24209351734187373       0.24144148153406236   0.23816259853651978
BERT(L11)       0.23780101536455783  0.24116252655452086       0.23842041488247545 0.2349979488477501
UniSentEnc      0.16984813537392474 	0.17011582363474143 0.16883505590379805 0.16603733206907578
GloVe           0.14713957888727855	    0.1477559756946128 0.1455671055893037 0.14305336437528837
}\tableone

\pgfplotstableread{
Model           Delay=-4     Delay=-2 Delay=0     Delay=2     Delay=4     Delay=6 Delay=8 Delay=10 E-Delay=-4  E-Delay=-2   E-Delay=0     E-Delay=2     E-Delay=4     E-Delay=6  E-Delay=8   E-Delay=10
GoogleLM(L0)	0.343575  	0.343588  	0.345213  	0.349875  	0.354602  	0.351002  	0.348563  	0.347392	0.027153  	0.026600  	0.025420  	0.023467  	0.022796  	0.022501  	0.023808  	0.023351
GoogleLM(L1)	0.271404  	0.272159  	0.275228  	0.281469  	0.285399  	0.281457  	0.277889  	0.276791	0.023117  	0.021729  	0.020574  	0.018105  	0.016330  	0.016301  	0.017568  	0.018093
ELMO(L0)	0.294963  	0.295180  	0.296787  	0.303175  	0.307938  	0.303565  	0.299885  	0.299688	0.020958  	0.020047  	0.019296  	0.016946  	0.015533  	0.015698  	0.016943  	0.016909
ELMO(L1)	0.260716  	0.261569  	0.264157  	0.271473  	0.275734  	0.271237  	0.267168  	0.266243	0.018838  	0.017824  	0.017161  	0.014641  	0.012948  	0.012482  	0.013321  	0.013094
BERT(L0)	0.234603  	0.234743  	0.234949  	0.237449  	0.238211  	0.235698  	0.234726  	0.234444	0.010340  	0.010156  	0.009943  	0.008865  	0.008127  	0.008616  	0.008924  	0.009544
BERT(L11)	0.227001  	0.227870  	0.230196  	0.234737  	0.235501  	0.232809  	0.232059  	0.232631	0.018067  	0.017030  	0.016473  	0.015192  	0.013735  	0.014085  	0.014836  	0.015307
UniSentEnc	0.171122  	0.173458  	0.175264  	0.176276  	0.175887  	0.174843  	0.174002  	0.173323	0.017186  	0.016040  	0.014790  	0.013591  	0.013008  	0.012054  	0.012568  	0.013297
GloVe	0.135831  	0.135860  	0.136625  	0.140097  	0.142728  	0.141485  	0.138522  	0.138188	0.008213  	0.007758  	0.008501  	0.008061  	0.007407  	0.007143  	0.007708  	0.008188
}\tablethree

\pgfplotstableread{
ContextLength	GoogleLM(L0)	GoogleLM(L1)	Elmo(L0)	Elmo(L1) BERT(L0)	BERT(L11) UniSentEnc GloVe
0 0.36785948515739875 0.3243572170126295 0.30361015369842403 0.25441500468956135 0.25442116569149564 0.21360143307364432 0.2055772657042046 0.14753044939469254
1 0.3775870285115762 0.29895023725970554 0.31632012736785614 0.282962040350316 0.23882216055801714 0.23780101536455783 0.16984813537392474 0.14713957888727855
2 0.37659338354472816 0.28308954757603166 0.3210871026121255 0.275706067339209 0.24900725248778247 0.23131494415636528 0.15794386557339396 0.14744829046043287
3 0.3745792792911261 0.292622481762721 0.3223460687650853 0.28496954400547025 0.246129252635169 0.22994967477292344 0.16480370947052697 0.14849229932707692
4	0.37357501417469624 0.29103490800839543 0.32213163144790846 0.282475760058574 0.24014165259155956 0.23747315435027377 0.15839703521564355 0.14787472202273994
5	0.37317128049610415 0.28820903731693165 0.32164493502887015 0.28140388554405327 0.24099159632584843 0.2369365054754422 0.16029371043689314 0.1480174606459751		
6	0.37291301573248503 0.2866577315053637 0.32160693636369586 0.28021130763440005 0.24066318475147253 0.24084276573219843 0.1602645620686417 0.1474464173609028
7	0.372790462061551 0.28490934672772106 0.3213892841771204 0.2794787464963043 0.23465552254496158 0.2477459231380783 0.14613117116287277 0.14750138734925847
}\tablefour

\pgfplotstableread{
ContextLength	GoogleLM(L0)	GoogleLM(L1)	Elmo(L0)	Elmo(L1) BERT(L0)	BERT(L11) UniSentEnc GloVe
0 0.3706315057195222 0.3277882568956974 0.30686394982620363 0.2583554959517297 0.2555259198736407 0.21575939350118656 0.2086599504842027 0.14802495656394055 
1 0.3812284259692499 0.3032681346108651 0.3193476497203331 0.2859980247689471 0.24209351734187373 0.24116252655452086 0.17011582363474143 0.1477559756946128 
2 0.3799278633379336 0.28530613281639733 0.3243471993849084 0.27891742782453177 0.25171537439695174 0.23412822615372628 0.15884179474753113 0.14809900980212853 
3 0.3781163300554886 0.2963615575640389 0.32608851796732896 0.2894604172370627 0.24790426524190956 0.2329646293950387 0.164358923044841 0.1491837121738585 
4 0.3770886611432551 0.2942203021010055 0.32573073721813955 0.2865415569952049 0.24326764782659832 0.24074774696786438 0.15793694450227191 0.14846487860055207 
5 0.37667252308758825 0.29123276505006823 0.32526061891306657 0.2854171138514021 0.24387975006417337 0.23966800338854188 0.15978966626605373 0.14864239310903937 
6 0.3763924665985593 0.28969692467274094 0.32521213294647827 0.28401472034225894 0.24273138252612794 0.24282295162409945 0.1603854424467102 0.1479948302544029 
7 0.37626034065067915 0.28771370838791055 0.32499717202546696 0.28344640475993205 0.236134665858372 0.24978913344213904 0.14782402297831695 0.1481433317933767 

}\tablethree

\pgfplotstableread{
ContextLength	GoogleLM(L0)	GoogleLM(L1)	Elmo(L0)	Elmo(L1) BERT(L0)	BERT(L11) UniSentEnc GloVe
0 0.3659771821936663 0.3255509632943318 0.3028033378852309 0.25411933068848047 0.2531846081670389 0.21303917873133835 0.20647817190282627 0.1458776480560079 
1 0.37658243835458194 0.2988682648688573 0.3136933609951579 0.27971861626651834 0.24144148153406236 0.23842041488247545 0.16883505590379805 0.1455671055893037 
2 0.3750283918375569 0.28144761968641974 0.3204254105466697 0.2751876985885664 0.25084917064503964 0.2342772986944816 0.15984307246353335 0.14578660383211217 
3 0.37305247218970217 0.29307370386560105 0.3219625487263893 0.2857922985319364 0.24644834054031567 0.23328458474783303 0.16452886503685424 0.14693993303945194 
4 0.37209670096504366 0.2909925251842502 0.321540500865803 0.28269514730240364 0.2432685707619128 0.2416580468956827 0.1575280691208771 0.14627940957155436 
5 0.3716105478060834 0.287445601074254 0.3211004012052522 0.28157920647625323 0.24395016474679404 0.23975971734070722 0.1586870902914993 0.1464479231112587 
6 0.3713126285284099 0.28591975545407067 0.32103439549531737 0.2799993846536648 0.24347640804034942 0.24346012026020458 0.15961854631058925 0.1457640597125771 
7 0.37117663743710516 0.2840764135000145 0.32080492086293183 0.2795145156300467 0.23795337582994303 0.25111988987508815 0.1496183468310896 0.14590812116887403 
	
}\tabletwo

\pgfplotstableread{
ContextLength	GoogleLM(L0)	GoogleLM(L1)	Elmo(L0)	Elmo(L1) BERT(L0)	BERT(L11) UniSentEnc GloVe
0 0.36281465922928857 0.32209732814561587 0.2979387980911101 0.24926513871233896 0.24910972793660227 0.2102707714397819 0.2028926635630494 0.1431780201676665 
1 0.3730588640988145 0.29388291720525195 0.3083983104688391 0.27412382969628624 0.23816259853651978 0.2349979488477501 0.16603733206907578 0.14305336437528837 
2 0.3713637045995286 0.2774218565937679 0.31548659303610815 0.26970270183539957 0.247889438843174 0.23037677803235615 0.15771045559484037 0.1431375124578671 
3 0.3693545835346882 0.2891622009185858 0.3168276427442114 0.27990619907627234 0.24375369002719635 0.2297786713702613 0.1634639747175373 0.14407726961730666 
4 0.3683833374355053 0.28696679879050146 0.3162808956823413 0.27685654485150135 0.24074498610669431 0.23861836032698802 0.15555894791054825 0.14357448340251416 
5 0.36788910077125186 0.28351202701241063 0.31590320418994366 0.2757648461657831 0.2408545518034694 0.2363294030995131 0.1574906319068835 0.14380630349028586 
6 0.3675832871258204 0.2818322347911898 0.31581222146332105 0.2740236785714414 0.24041894296707947 0.24071786644142718 0.15814165663646262 0.1431246781706338 
7 0.36743891482389424 0.2800209511371097 0.3156089116098857 0.27358510001707387 0.23644052720630407 0.24818954487966646 0.14851595703957496 0.14324503340157102 

}\tableone

\paragraph{Different Segments of the Story}
If during training the models are only trained on full sentences, it might be the case that the quality of their representations, when given complete sentences, is significantly better than when provided with incomplete sentences. 
On the other hand, the representation of sentences in the brain might also be more reliable when the full sentence is read.
To take this into account, we look at the similarities of each of the models with brain representations, only at the steps in the story where an end of a sentence token is reached.  Figure~\ref{fig:perf_in_story_end} presents the results. We see that in this case, the similarity of all the models with brain representations increases slightly, but this could be because of the reduced dimensionality of the similarity matrix, and we see that the general patterns stay similar.

In Figure \ref{fig:perf_in_story_char} we observe  that at the story segments where a name of a character is mentioned, the patterns of similarities change a bit, e.g. the last layer of BERT is less similar to the brain representations compared the first layer of BERT, when an intermediate amount of context is provided to the model. This finding is difficult to interpret, but warrants further research.
\usetikzlibrary{patterns}

\pgfplotstableread{
ContextLength	GoogleLM(L0)	GoogleLM(L1)	Elmo(L0)	Elmo(L1) BERT(L0)	BERT(L02)	UnisentEnc GloVe
0 0.542410121678004 0.5576931444463515 0.5326929732562049 0.4889386246863532 0.48643226209723345 0.4048097836424608 0.42689130965471567 0.313158455090394 
1 0.5444931928190873 0.5246493093518476 0.540577925108487 0.5155957216430542 0.480379649124004 0.48148650217649025 0.3508017947865487 0.31324123908645224 
2 0.5552410373145411 0.49055796809065255 0.5413012876921595 0.5152822112413773 0.47881260349423294 0.47516454847907397 0.3251441846290429 0.31301764390750086 
3 0.5620830298006751 0.5210586858814854 0.5427748058464428 0.515727413488663 0.47027966693765 0.47468523217687125 0.34055695880558534 0.31304850951063923 
4 0.5631041784534281 0.5218848035038499 0.5431272168250446 0.514683924769548 0.4591817728756061 0.49144940898808087 0.33066959074142355 0.3131257802734101 
5 0.5611242595201145 0.5139193839209287 0.5431540482705268 0.5150343523100676 0.45942169322326565 0.48370416177988973 0.33675101361713883 0.31303039405722216 
6 0.5613496815214924 0.513182162348988 0.5435821085883478 0.5144885693099224 0.4608993360677559 0.5014055963242509 0.3221575623771875 0.3130079307942025 
7 0.561709198464625 0.5119313368943519 0.5433361240713441 0.5130347439397885 0.46415247765568124 0.5118067938116463 0.31113273137703035 0.31331652043282227 
		
}\tableone

\pgfplotstableread{
ContextLength	GoogleLM(L0)	GoogleLM(L1)	Elmo(L0)	Elmo(L1) BERT(L0)	BERT(L02)	UnisentEnc GloVe
0 0.5619129947850562 0.5529861791711829 0.5202205669708354 0.4691405928860034 0.47381185980465634 0.4058157602257585 0.41580827306873486 0.3003986823502194 
1 0.5667846622335131 0.47112433862518704 0.512623072046853 0.4931704262357439 0.45597401489541733 0.48456400340540295 0.3814697136010058 0.3006041169555233 
2 0.5697597924652233 0.47948483414151455 0.5284007503367911 0.4934974229745595 0.4746553662329501 0.4745297118074826 0.38578970410516267 0.30076916337411064 
3 0.5600546944435963 0.5029673669973327 0.5200786789677869 0.4992918031983199 0.45111913737232767 0.46593792497231135 0.3862617954783255 0.30025794150186347 
4 0.557899513277548 0.5084114165766285 0.5200930197333619 0.5033303128000408 0.4578148928194422 0.4849732149719773 0.3906211762148051 0.30061034944756293 
5 0.5574814887265287 0.5123972318199784 0.5187071824948883 0.5004626510613938 0.4466917641530597 0.4677784316286102 0.39963122986368294 0.30040825286927786 
6 0.5564506524590676 0.5104980836203767 0.5183734562580401 0.499098143215182 0.4496091394050873 0.47462949776377616 0.397711116474461 0.3006247270312813 
7 0.5560053545037964 0.5111049696388571 0.5181708339090464 0.49913604159173375 0.45297777982017007 0.48694971711132506 0.39268714755427614 0.30110678768058524 

}\tabletwo

\pgfplotstableread{
ContextLength	GoogleLM(L0)	GoogleLM(L1)	Elmo(L0)	Elmo(L1) BERT(L0)	BERT(L02)	UnisentEnc GloVe
0 0.654096732389038 0.6914380207188486 0.6398028262116937 0.618530450699385 0.6129620771459395 0.618187280551984 0.5320342345707494 0.4765258703854003 
1 0.6760982055248292 0.6391619844026447 0.653240147387066 0.6797388051032935 0.619330336446396 0.6105267710174096 0.5261820479946235 0.4765258703854003 
2 0.6941708833346443 0.6289391631015373 0.6673285849497681 0.6554926127388156 0.6297129570332805 0.5700189504295161 0.504443998312815 0.4765258703854003 
3 0.6897725274658437 0.6624459049715996 0.6685475201266853 0.6869323571406125 0.6455514179846176 0.5568948626976293 0.46084221085156685 0.4765258703854003 
4 0.6921270659119501 0.6647825590489913 0.6728575256685698 0.6889121358998541 0.6175189774537017 0.574985424359009 0.48948442178903406 0.4765258703854003 
5 0.6918722460069825 0.6572360304364995 0.6709444333356701 0.6833506864872316 0.6388435486273885 0.6162622171872479 0.4864289633767189 0.4765258703854003 
6 0.6921889173795881 0.6563665667316606 0.672132034068829 0.6822946208255709 0.6323542042026373 0.6338044849175941 0.4612590019713505 0.4765258703854003 
7 0.6919947593330376 0.6555140831785622 0.6715643340817374 0.6811835767399029 0.6393441815378309 0.6453934637016894 0.4362916641162602 0.4765258703854003 
}\tablethree

\begin{figure*}
\centering
\begin{subfigure}[b]{\columnwidth}
\begin{tikzpicture}
\begin{axis}[
width=\columnwidth+0.9cm, 
height=4cm,
xmajorgrids,
minor tick num=1,
xlabel={\texttt{Context Length}},
xtick=data,
ymax=0.2, 
ymax=0.6,
tick label style = {font=\fontsize{5}{6}\selectfont},
xtick={0,1,2,3,4,5,6,7},
xticklabels={0,1,2,3,4,5,6,7},
label style = {font=\fontsize{6}{7}\selectfont, yshift=0.5ex},
legend style={at={(2.0,1.20),font=\fontsize{5}{6}\selectfont}
  ,legend columns=8
  },
]
\addplot [thick, NavyBlue, dashed, mark=square, mark options={scale=0.5, solid}] table[x index=0, y index=1]{\tableone};

\addplot [thick, Cyan, dashed, mark=square, mark options={scale=0.5, solid}] table[x index=0,y index=2] {\tableone};

\addplot [thick, Mulberry, dashed , mark=triangle, mark options={scale=0.5, solid}] table[x index=0, y index=3] {\tableone};

\addplot [thick, Orchid, dashed, mark=triangle,mark options={scale=0.5, solid}] table[x index=0, y index=4] {\tableone};

\addplot [thick, RedOrange, dashed, mark=o,mark options={scale=0.5, solid}] table[x index=0, y index=5]{\tableone};

\addplot [thick, BurntOrange, dashed, mark=o,mark options={scale=0.5, solid}] table[x index=0, y index=6] {\tableone};

\addplot [thick, PineGreen, dashed, mark=+,mark options={scale=0.5, solid}] table[x index=0, y index=7] {\tableone};

\addplot [thick, YellowGreen, dashed, mark=+,mark options={scale=0.5, solid}] table[x index=0, y index=8] {\tableone};


\legend{GoogleLM(L0), GoogleLM(L0), Elmo(L0), Elmo(L0), BERT(L0), BERT(L01), UniSentEnc, GloVe}
\end{axis}
\end{tikzpicture}
    \caption{\fontsize{7}{8}\selectfont{Complete sentences}
    \label{fig:perf_in_story_end}}
    \end{subfigure}
~
\begin{subfigure}[b]{\columnwidth}
\begin{tikzpicture}
\begin{axis}[
width=\columnwidth+0.9cm, 
height=4cm,
xmajorgrids,
minor tick num=1,
xlabel={\texttt{Context Length}},
xtick=data,
ymax=0.2, 
ymax=0.75,
tick label style = {font=\fontsize{5}{6}\selectfont},
xtick={0,1,2,3,4,5,6,7},
xticklabels={0,1,2,3,4,5,6,7},
label style = {font=\fontsize{6}{7}\selectfont, yshift=0.5ex},
]
\addplot [thick, NavyBlue, dashed, mark=square, mark options={scale=0.5, solid}] table[x index=0, y index=1]{\tablethree};

\addplot [thick, Cyan, dashed, mark=square, mark options={scale=0.5, solid}] table[x index=0, y index=2] {\tablethree};

\addplot [thick, Mulberry, dashed , mark=triangle,mark options={scale=0.5, solid} ] table[x index=0, y index=3] {\tablethree};

\addplot [thick, Orchid, dashed, mark=triangle, mark options={scale=0.5, solid}] table[x index=0, y index=4] {\tablethree};

\addplot [thick, RedOrange, dashed, mark=o, mark options={scale=0.5, solid}] table[x index=0, y index=5]{\tablethree};

\addplot [thick, BurntOrange, dashed, mark=o, mark options={scale=0.5, solid}] table[x index=0, y index=6] {\tablethree};

\addplot [thick, PineGreen, dashed, mark=+, mark options={scale=0.5, solid}] table[x index=0, y index=7] {\tablethree};

\addplot [thick, YellowGreen, dashed, mark=x, mark options={scale=0.5, solid}] table[x index=0, y index=8] {\tablethree};


\end{axis}
\end{tikzpicture}
    \caption{\fontsize{7}{8}\selectfont{Mentions of story character}
    \label{fig:perf_in_story_char}}
    \end{subfigure}
\caption{Similarity of the computational representations with brain representations at different segments of the story. \label{fig:perf_in_story}}
\end{figure*}
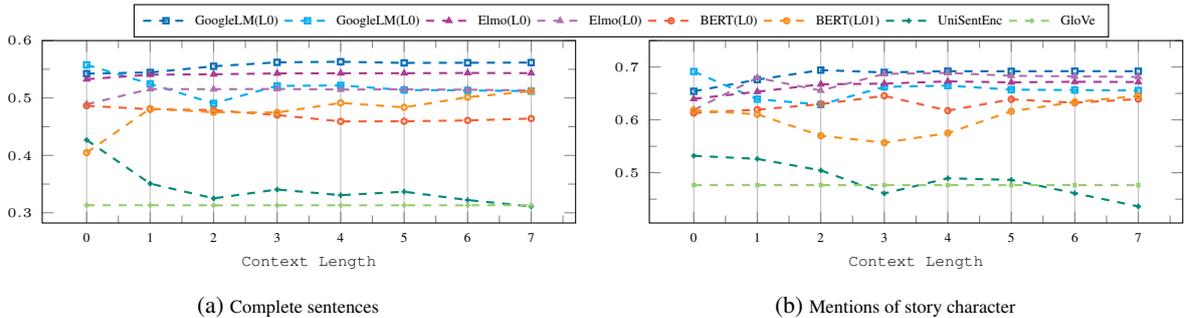

\begin{figure}
    \centering
    \begin{subfigure}[b]{\columnwidth}
    \includegraphics[width=\columnwidth]{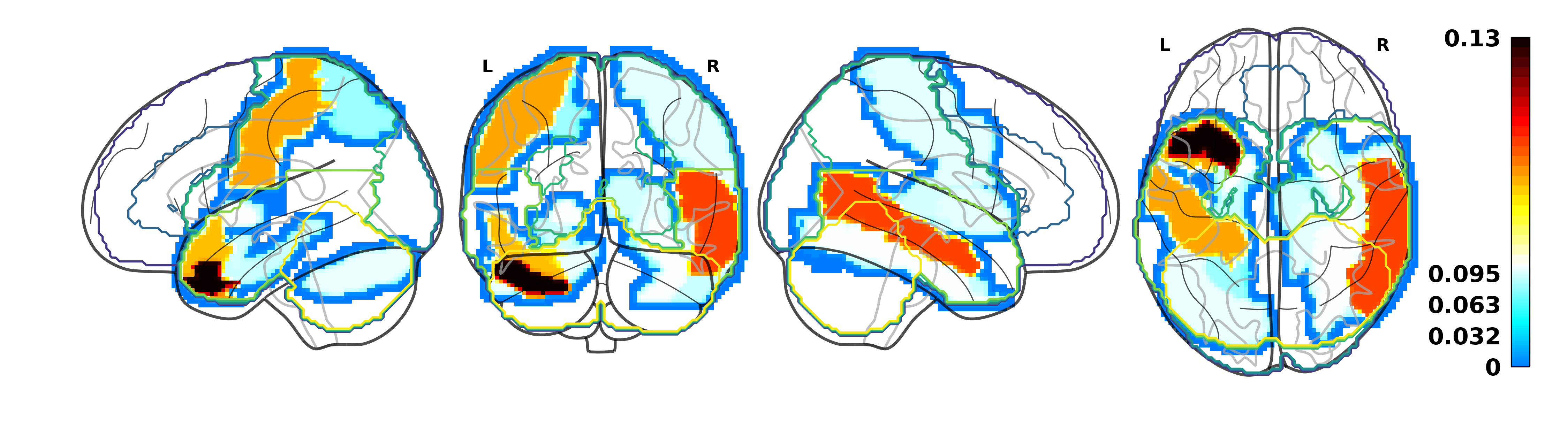}
      \vspace{-16pt}
    \caption{\fontsize{6}{7}\selectfont{GoogleLM (L0)}}
    \end{subfigure}
    \begin{subfigure}[b]{\columnwidth}
    \includegraphics[width=\columnwidth]{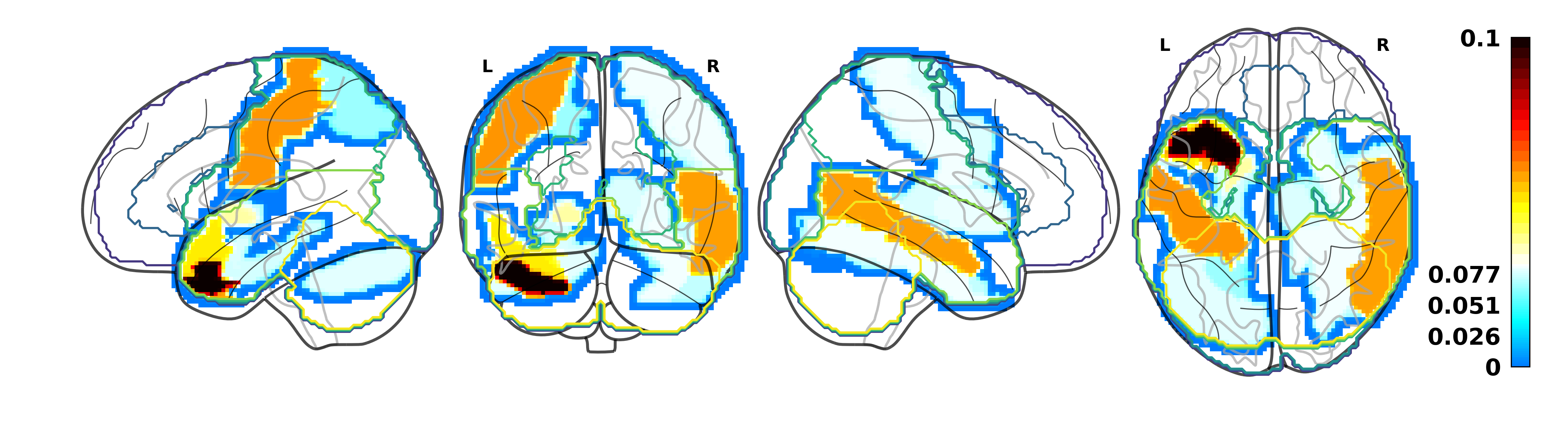}
      \vspace{-16pt}
    \caption{\fontsize{6}{7}\selectfont{GoogleLM (L1)}}
    \end{subfigure}
    \begin{subfigure}[b]{\columnwidth}
    \includegraphics[width=\columnwidth]{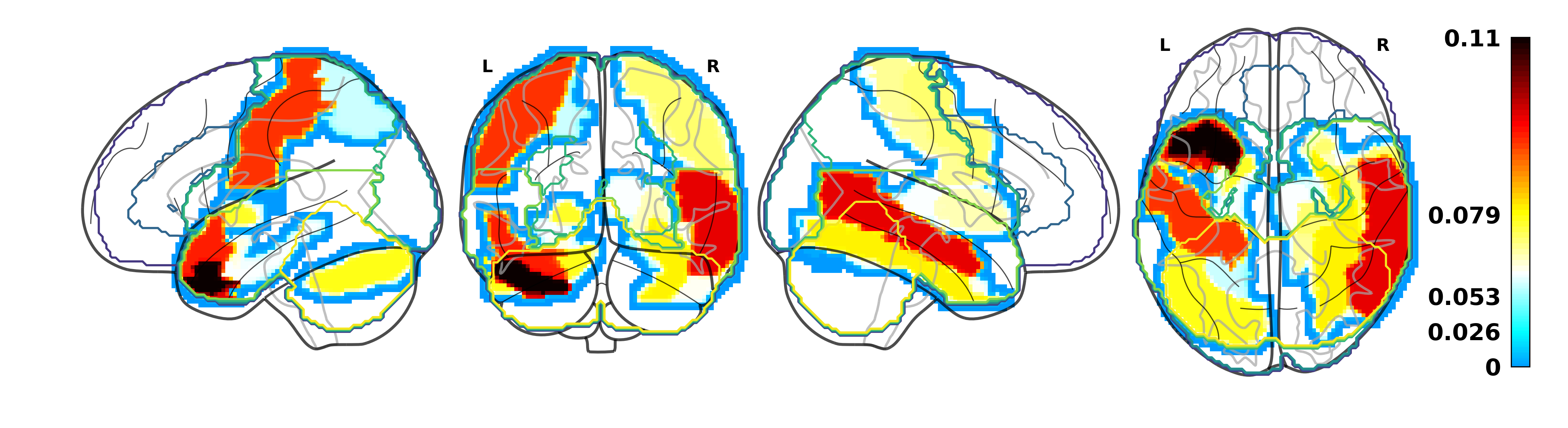}
      \vspace{-16pt}
    \caption{\fontsize{6}{7}\selectfont{ELMO (L0)}}
    \end{subfigure}
    \begin{subfigure}[b]{\columnwidth}
    \includegraphics[width=\columnwidth]{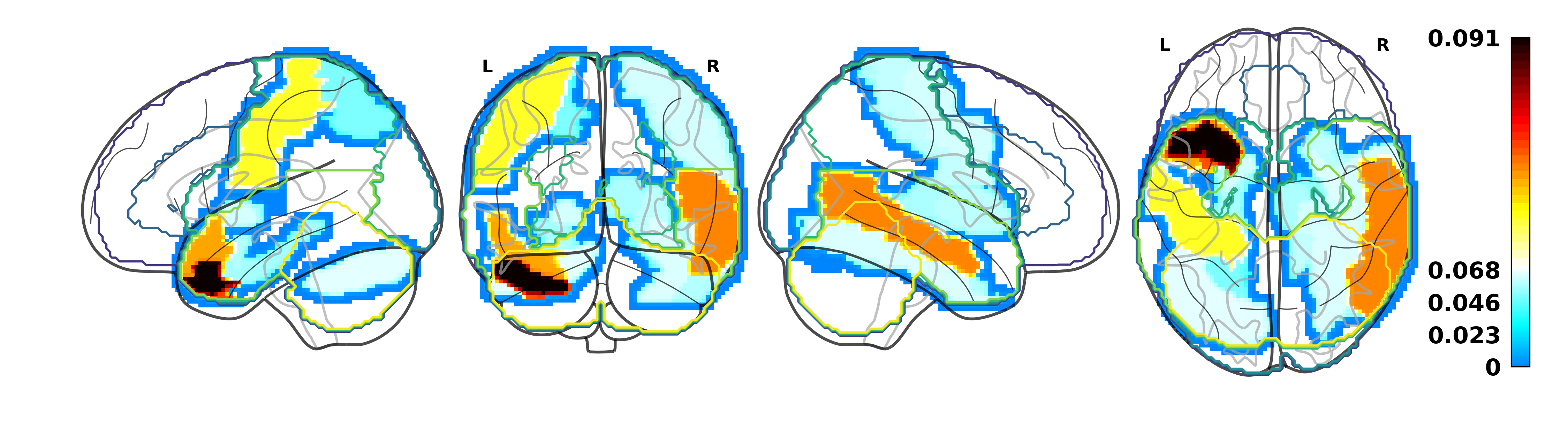}
      \vspace{-16pt}
    \caption{\fontsize{6}{7}\selectfont{ELMO (L1)}}
    \end{subfigure}
    \begin{subfigure}[b]{\columnwidth}
    \includegraphics[width=\columnwidth]{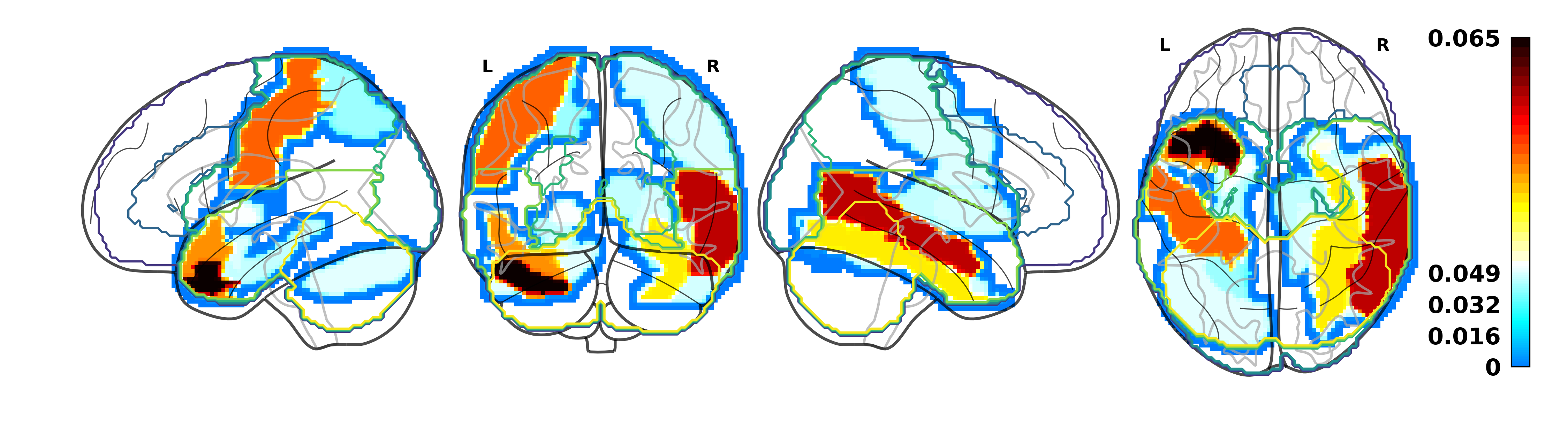}
      \vspace{-16pt}
    \caption{\fontsize{6}{7}\selectfont{UniSentEnc}}
    \begin{subfigure}[b]{\columnwidth}
    \includegraphics[width=\columnwidth]{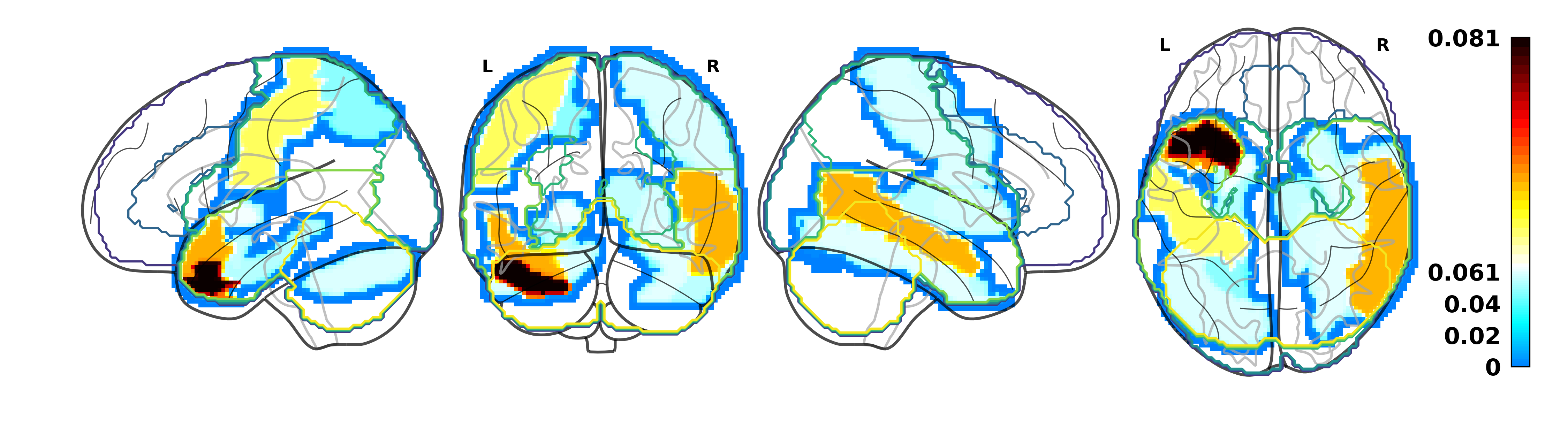}
      \vspace{-16pt}
    \caption{\fontsize{6}{7}\selectfont{BERT (L0)}}
    \end{subfigure}
    \begin{subfigure}[b]{\columnwidth}
    \includegraphics[width=\columnwidth]{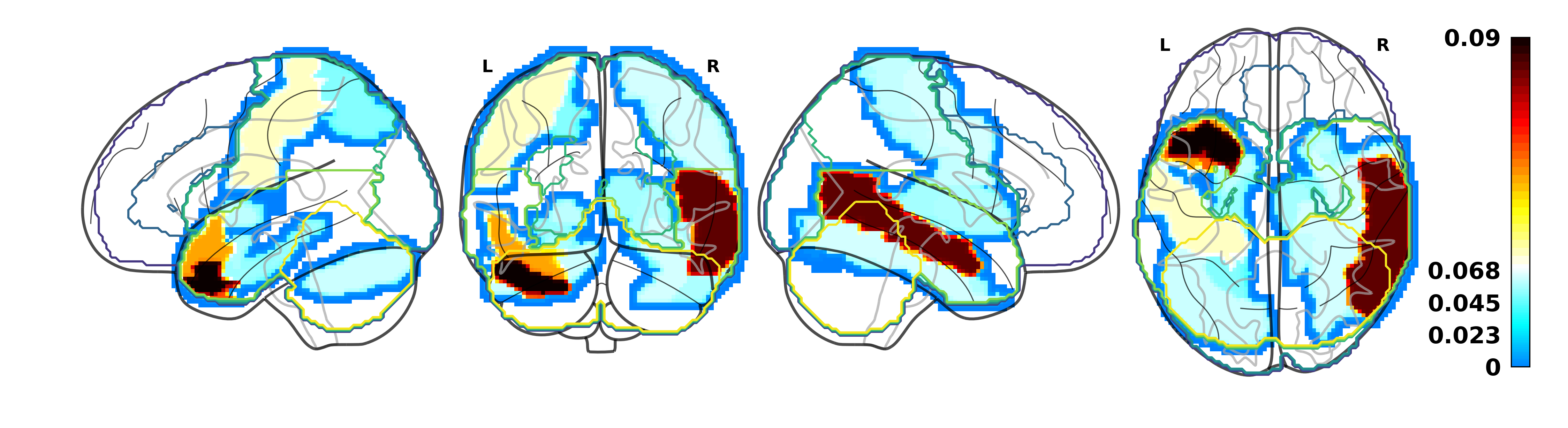}
      \vspace{-16pt}
    \caption{\fontsize{6}{7}\selectfont{BERT (L11)}}
    \end{subfigure}
    \end{subfigure}
    \begin{subfigure}[b]{\columnwidth}
    \includegraphics[width=\columnwidth]{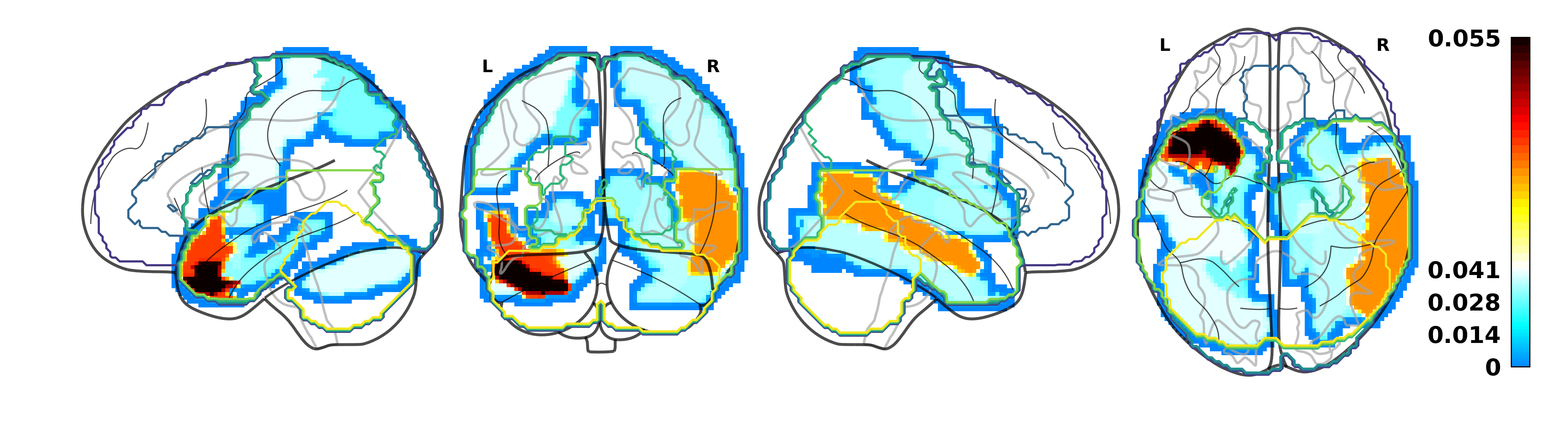}
      \vspace{-16pt}
    \caption{\fontsize{6}{7}\selectfont{GloVe}}
    \end{subfigure}
    \vspace{-15pt}
    \caption{RSA of representations learned at different layers of different models with  representations at different regions of Subject4's brain which is chosen randomly (the code accompanying this paper can be used to generate the plots for the other subjects). In order to emphasize the difference of the similarity of each model with different brain regions, the color bar is scaled independently for each model. The darkest region for all models is the Left Anterior Temporal Lobe. 
    \label{fig:brain_plots}}
\end{figure}

\paragraph{Different Regions of the Brain}
We looked at the similarity scores of the computational representations with the representations at different regions of the brain.  This is illustrated in Figure \ref{fig:brain_plots} for subject 4 as an example.
We observe that the patterns of RSA of different models are very similar across different brain regions, i.e. the scores scale for all regions almost similarly across different models. Despite the low correlations between the models and the brain activation, we find that all the models are consistently best aligned with the regions in the Left Anterior Temporal Lobe (LATL). This region is known for semantic and sometimes syntactic processing of language~\cite{westerlund2014role, bemis2011simple, leffel2014restrictive}. We also find some correlation with the Left Parietal Lobe, which is not known to be responsible for language processing.
We also computed the average RSA between different brain regions for the eight subjects, both within and across subjects, and find that the different regions of a single brain are more similar ($RSA=0.4$) than the same regions of different brains ($RSA=0.12$). 
These are counter-intuitive findings that warrant further investigation. If brain functions involved in story comprehension are spatially localized and brains are organized similarly across individuals, we would expect the same regions from different subjects to be more similar than different regions from the same subject.

\usetikzlibrary{patterns}

\pgfplotstableread{
Delay Sub(1) Sub(2) Sub(3) Sub(4) Sub(5) Sub(6) Sub(7) Sub(8) 
0 0.010123240985822846 0.007626132635401434 0.006602685688054888 0.005410089633506643 0.005376826018021724 0.006016002004173487 0.005848763224629766 0.005141107224453573
2 0.008768918215278365 0.007533157688462827 0.00627454721353235 0.005547282711523799 0.006157467046340237 0.005811082391647682 0.005354684684741456 0.005052911336392185
4 0.00749456271096699 0.007331906474880773 0.005953117563751414 0.00570727219315445 0.006676698396707814 0.005796520973628316 0.005745771991700371 0.005455127347783362
6 0.007225029993934207 0.007149753119834568 0.006084309026267137 0.005813333581397601 0.00624704701543255 0.006529045722786703 0.005784483755459258 0.005686269894688928
8 0.007118475704526395 0.007266528631129782 0.006073393211853767 0.005917260145354347 0.005882428337453588 0.006395702545609616 0.005691780953921394 0.005684495918481636
10 0.006927410641312498 0.006539033571898251 0.005626996580996663 0.005844590397056565 0.00596101474806332 0.006095420051964673 0.005965465772810485 0.006050633995100779
}\tableone

\pgfplotstableread{
Delay Sub(1) Sub(2) Sub(3) Sub(4) Sub(5) Sub(6) Sub(7) Sub(8) 
0 0.005800858982075691 0.004718704824233949 0.0050400414700629 0.005260566115488002 0.004718453020584185 0.0052432225087269135 0.004631637381281362 0.0058901222949995224
2 0.006193552189062229 0.004905933301959181 0.005411662529897859 0.006679798798192277 0.0046875790969459485 0.005466726477440925 0.004864537381759904 0.006239262674564039
4 0.007966856512450576 0.005083263011298189 0.006450870978794376 0.008531591496756585 0.005774093685424836 0.006096359847837907 0.005897679610664369 0.006885460873412241
6 0.009379972139781534 0.005151963871893428 0.007041500969316911 0.008380988224952896 0.006500631081410357 0.0071267380575678885 0.006300961540353589 0.006462379449049427
8 0.007516832204338131 0.004907650747388748 0.006198381704401873 0.006822353878981058 0.005245606333643016 0.007273420476040682 0.005874622046656314 0.00498084524282702
10 0.005878021318936583 0.004788436971724097 0.005599011404952808 0.005651517640916953 0.004882632189412955 0.006302884115817614 0.0050695892182220925 0.004515837992019683
}\tabletwo

\pgfplotstableread{
Delay Sub(1) Sub(2) Sub(3) Sub(4) Sub(5) Sub(6) Sub(7) Sub(8) 
0 0.006899579293386193 0.006602527580149342 0.006324983497427189 0.0076093689465322745 0.007107284123940183 0.010103632952094155 0.006721753097558347 0.007454056607056309
2 0.007232395284836324 0.006735301665145577 0.006816658064076664 0.010100946012893658 0.006865202618448113 0.010140727369413266 0.0073197247087236905 0.008223333124232984
4 0.008225626379521456 0.00680657327711684 0.008410428298845113 0.013362906714640133 0.008109621562142455 0.011957205662475573 0.00829194682591547 0.008227008672935604
6 0.01002424816919259 0.006378512394081881 0.009292333724601004 0.01314933956351913 0.0077692175249188256 0.012623982374464993 0.009198848456015957 0.0075997574511961485
8 0.00916683571410512 0.006210511667967267 0.007482859621447763 0.01091888215494663 0.0069925065313638235 0.01191106583865173 0.007694421213868707 0.006328844999818284
10 0.007794976527802977 0.006013704802286718 0.007056494824548815 0.008762920526719009 0.0060163251368343384 0.010053423420905094 0.006518315766588456 0.0056730346350781224
}\tablethree

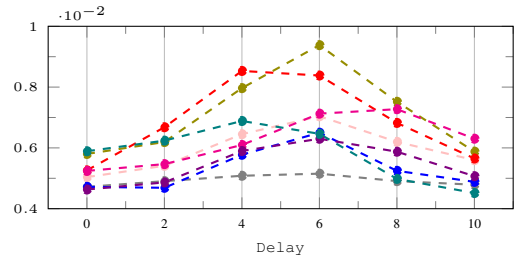
\begin{figure}[t!]
\begin{tikzpicture}
\begin{axis}[
width= \columnwidth, 
height=4cm,
xmajorgrids,
minor tick num=1,
xlabel={\texttt{Delay}},
xtick=data,
ymin=0.004, 
ymax=0.01,
tick label style = {font=\fontsize{5}{6}\selectfont},
xtick={0,2,4,6,8,10},
xticklabels={0,2,4,6,8,10},
label style = {font=\fontsize{6}{7}\selectfont, yshift=0.5ex},
]
\addplot [thick, olive, dashed, mark=*,  mark size=1.5pt] table[x index=0, y index=1]{\tabletwo};

\addplot [thick, gray, dashed, mark=*,  mark size=1.5pt] table[x index=0,y index=2] {\tabletwo};

\addplot [thick, pink, dashed , mark=*,  mark size=1.5pt] table[x index=0,y index=3] {\tabletwo};

\addplot [thick, red, dashed, mark=*,  mark size=1.5pt] table[x index=0,y index=4] {\tabletwo};

\addplot [thick, blue, dashed, mark=*,  mark size=1.5pt] table[x index=0, y index=5]{\tabletwo};

\addplot [thick, magenta, dashed, mark=*,  mark size=1.5pt] table[x index=0,y index=6] {\tabletwo};

\addplot [thick, violet, dashed , mark=*,  mark size=1.5pt] table[x index=0,y index=7] {\tabletwo};

\addplot [thick, teal, dashed, mark=*,  mark size=1.5pt] table[x index=0,y index=8] {\tabletwo};


\end{axis}
\end{tikzpicture}
\caption{Predictive power of representations learned by Google LM (L0\_Cinf) for brain representations in terms of explained variance (each subject in a different color).
\label{fig:predict-delay}}
\vspace{-10pt}
\end{figure}

\paragraph{Predictive Approach}
Besides, RSA, we can use a predictive approach to see which regions of the brain are more predictable, given the representations from a computational model.
In the predictive approach, we train a linear regression model to predict the brain activity patterns at different steps of the story. This way, we can obtain more fine-grained insights into which parts of the model contribute more to which regions in the brain.

In Figure \ref{fig:predict-delay}, we show the results of using representations obtained from GoogleLM($L0$) to predict brain activity patterns of different subjects. Similar to the results we obtained from RSA, the effect of hemodynamic response delay is clearly visible here. One of the difficulties of employing a predictive approach is to train a regression model for such high dimensions and with so little data. Hence, if the performance of the prediction is low, it is hard to tell if it is because we are not able to train a good regression model or because there is no correlation between the two models.  
To overcome this challenge, one solution could be to first use RSA to reduce the search space and then employ predictive modelling to gain more fine-grained insights. We postpone further analysis with the predictive approach to future studies.


\section{Discussion and Conclusion}
\label{sec:discussion}
In this paper, we employ a representational similarity metric to
compare the representations from the language encoding models with the brain activity patterns, i.e. measure the alignment between the brain activation patterns and activations of the internal state of the models.
The main advantage of RSA is that it treats both the brain and the model as a blackbox; it does not need to know how brains or models represent objects, words or sentences, but only how similar representations are to each other. For $N$ stimuli considered, the analysis only compares $\frac{1}{2}N(N-1)$ pairs of pairwise similarities (assuming similarities are symmetric), regardless of the dimensionality of two representational spaces. This bottleneck brings many advantages including computational efficiency, reuse of the similarity matrices in multiple comparisons, and not having to worry about how to map representations of very different nature to each other. It also brings important limitations and inevitable information loss, e.g. standard RSA, assumes all features of the representational spaces to have equal contributions.

One of our contributions in this paper is the introduction of ReStA, which uses RSA to measure the stability of the representations from the models when an input condition such as context length is changed. Comparing the representational similarity of different layers of different models, we find that both architectural differences and different training objectives have a noticeable impact on the representations learned by the models and the way they change under different conditions. We see a clear difference in the sensitivity to context size between $L0$ and $L1$ in the LSTM based models. This means, in line with results from previous work using different methods~\citep[e.g.,][]{giulianelli2018under}, that the $L1$ component integrates information over time steps while $L0$ does not.

Using brain data to evaluate the representations learned at different layers of each of the language encoding models, we find that layers of the LSTM based models achieve higher similarity score with brain data compared to single word representation models like GloVe and the Transformer based models. This observation could show that the learning biases of the LSTM based language models are closer to what happens in the human brain. Zooming into the results, we see that while changing the conditions of the inputs to the models has a significant impact on the representations they compute and their performance on NLP tasks~\cite{khandelwal2018sharp}, these changes do not get reflected in their alignment with the brain representations.

Finally, evaluating computational models of language processing with brain imaging data for a task such as ``story reading'' is hard, because of the inherent issues in the brain data and also the complexity of the task~\cite{beinborn2019robust}. 
Both the RSA framework and the predictive modelling approach make it possible to make a bridge between these black boxes, neural network models for language processing on the one hand and the human brain on the other. And while each of these approaches has its benefits and limitations, they might provide us with complementary information. Hence, it is invaluable to look at both of them. 

In our experiments, we observe more similarities between representations learned by some architectures and brain representations. However, caution is required when interpreting these results, as the representational similarity between all models and the brain images remains very low. We plan to perform further analysis on various (bigger) datasets to get a better interpretation of what is happening in both the brain and these computational models.


\section{Acknowledgement}
We thank Dieuwke Hupkes, Arnold Kochari, the Language in Interaction BQ1 team, and the anonymous reviewers for useful comments on the research described here and earlier versions of this paper. 
The work presented here was funded by the Netherlands Organization for Scientific Research (NWO), through a Gravitation Grant
024.001.006 to the Language in Interaction Consortium.

\newpage

\bibliography{main}
\bibliographystyle{acl_natbib}

\clearpage
\onecolumn
\section{Supplementary Material}
\subsection{Preprocessing Brain Images}
Besides the cognitive process of interest,  other factors like the physiological processes in the bodies of the human subjects or technical features of the MRI-machine and scanning environment may influence the fMRI measurements. An important issue is therefore how to preprocess the data to filter out those irrelevant effects adequately. 
\paragraph{Detrending.}
We normalise the brain activations in two steps: we scale the activation values by subtracting the per-voxel mean activation. We also experiment with a more elaborate preprocessing procedure, implemented in the \texttt{nilearn.signal.clean} Python library. Detrending is a popular strategy in cognitive neuroscience \cite{abraham2014machine}, that removes the linear trend, applies a high pass filtering with $0.005$ Hz, and standardises the vectors. 
\paragraph{Voxel selection.}
To reduce the noise and remove the voxels which their activation is not related to the story reading task, we apply two steps for selecting the voxels.
In the first step, we remove all the constant voxels. These are the brain regions in which the activation does not change at all during the scanning experiment.
Next, we compare the similarity of different regions of the brain for all eight subjects and select those regions that their activations over the different segments of the story are most similar among the different subjects. To do this, we rank the regions based on the average of the similarity scores and then selected the top 16 regions. 
After applying this voxel selection strategy, we have approximately 10000 voxels for each subject.

In our experiments, we do not model the spatial dependency of the voxels. Thus, after the preprocessing steps, we flatten the 3D fMRI images into vectors with the size of the total number of the voxels.

\clearpage

\subsection{Representational Similarity Across Different Layers of Different Models}
\begin{figure*}[h]
    \centering
    \includegraphics[width=\textwidth, clip, trim=5mm 4mm 5mm 5mm]{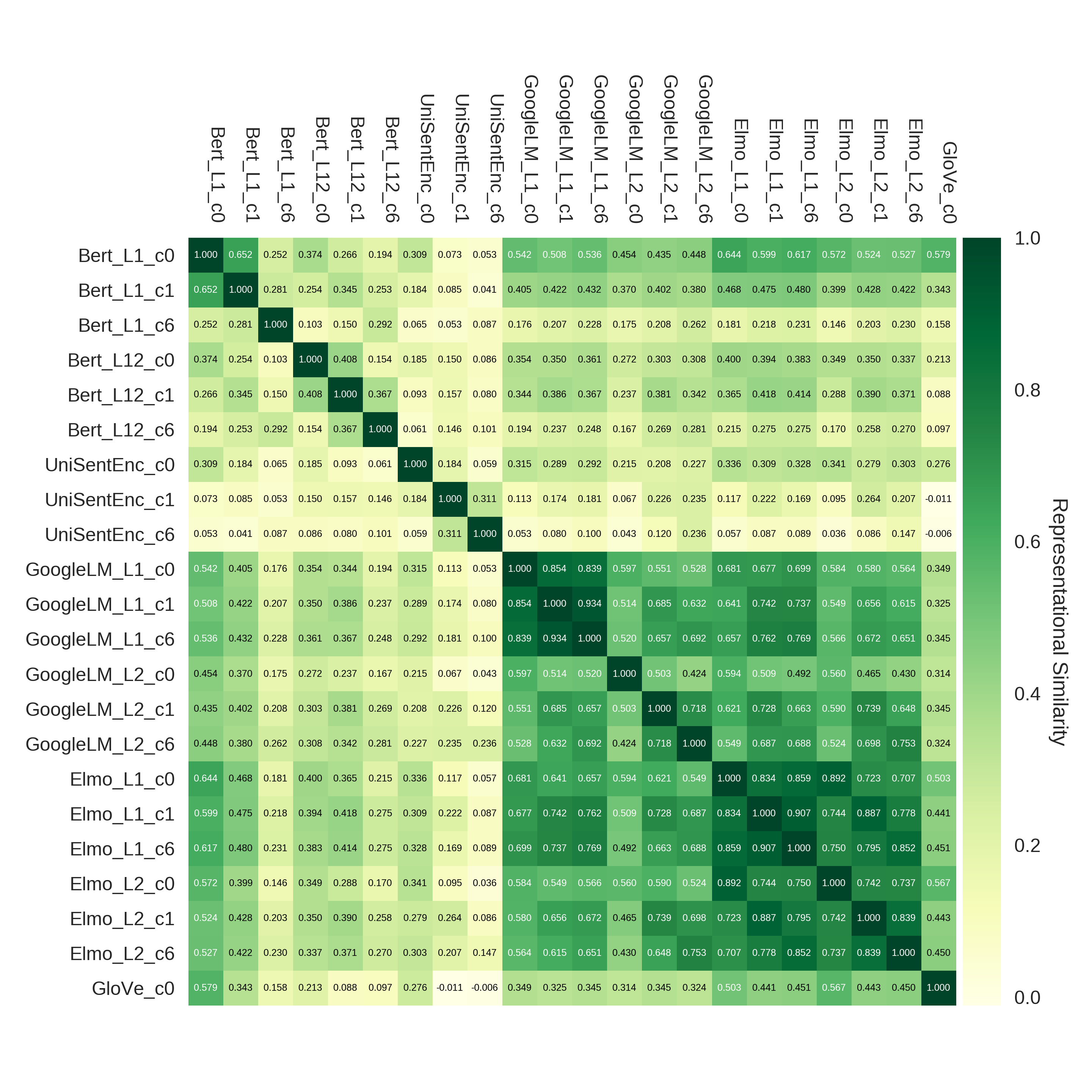}
    \caption{RSA of different layers of different models for different context length. In this plot, for example \texttt{ELMO\_0\_c1} means representation from layer 1 of ELMO, when the context length is 1 sentences.
    \label{fig:all_lm_sim}}
\end{figure*}

\end{document}